\documentclass[10pt,twocolumn,letterpaper]{article}

\usepackage{cvpr}              
\definecolor{cvprblue}{rgb}{0.21,0.49,0.74}
\usepackage[pagebackref,breaklinks,colorlinks,allcolors=cvprblue]{hyperref}
\usepackage{multirow}
\usepackage[table]{xcolor}
\usepackage{titletoc}

\usepackage[accsupp]{axessibility} 

\def\paperID{16679} 
\def\confName{CVPR}
\def\confYear{2026}

\def\netName{ForeSight}  

\title{See Tomorrow, Act Today: Foresight-Driven Autonomous Driving}

\author{
  Bozhou Zhang$^{1,2}$\thanks{Equal contribution, $^\ddagger$Corresponding author (\url{lizhangfd@fudan.edu.cn}).}
  ,
  Nan Song$^{1,2*}$
  ,
  Yuang Wang$^{1}$
  ,
  Jiankang Deng$^{3}$
  ,
  Xiatian Zhu$^{4}$
  ,
  Li Zhang$^{1,2\ddagger}$
  \vspace{.6em} 
  \\
  $^{1}$School of Data Science, Fudan University
  \quad 
  $^{2}$Shanghai Innovation Institute
  \vspace{.3em} 
  \\
  $^{3}$Imperial College London
  \quad 
  $^{4}$University of Surrey
  \vspace{.6em} 
  \\
  \url{https://github.com/LogosRoboticsGroup/ForeSight}
}

\begin{document}

\maketitle
\begin{abstract}

Current end-to-end autonomous driving planners are fundamentally reactive: they condition on historical and present observations to predict future actions. We argue that autonomous agents should instead imagine future scenes before deciding, just as human drivers mentally simulate ``what will happen next" before acting.
We introduce \textbf{\netName{}}, a foundation world model centric planning framework that reframes autonomous driving as anticipatory decision-making. Rather than treating world models as auxiliary components, \netName{} makes future scene imagination the primary driver of action prediction. Our approach operates in two stages: (1) generating plausible future visual worlds via a pretrained world model, and (2) planning actions conditioned on these imagined futures.
This paradigm shift from ``what should I do now?" to ``what will happen, and how should I respond?" enables genuinely anticipatory rather than reactive planning. By grounding decisions in anticipated contexts rather than present observations alone, \netName{} navigates dynamic, interactive scenarios more effectively.
Extensive experiments on NAVSIM and nuScenes demonstrate that explicit future imagination significantly outperforms previous state-of-the-art alternatives, validating our foresight-driven approach.

\end{abstract}
    
\section{Introduction}
\label{sec:intro}

\begin{figure}[t!]
\centering
\includegraphics[width=0.47\textwidth]{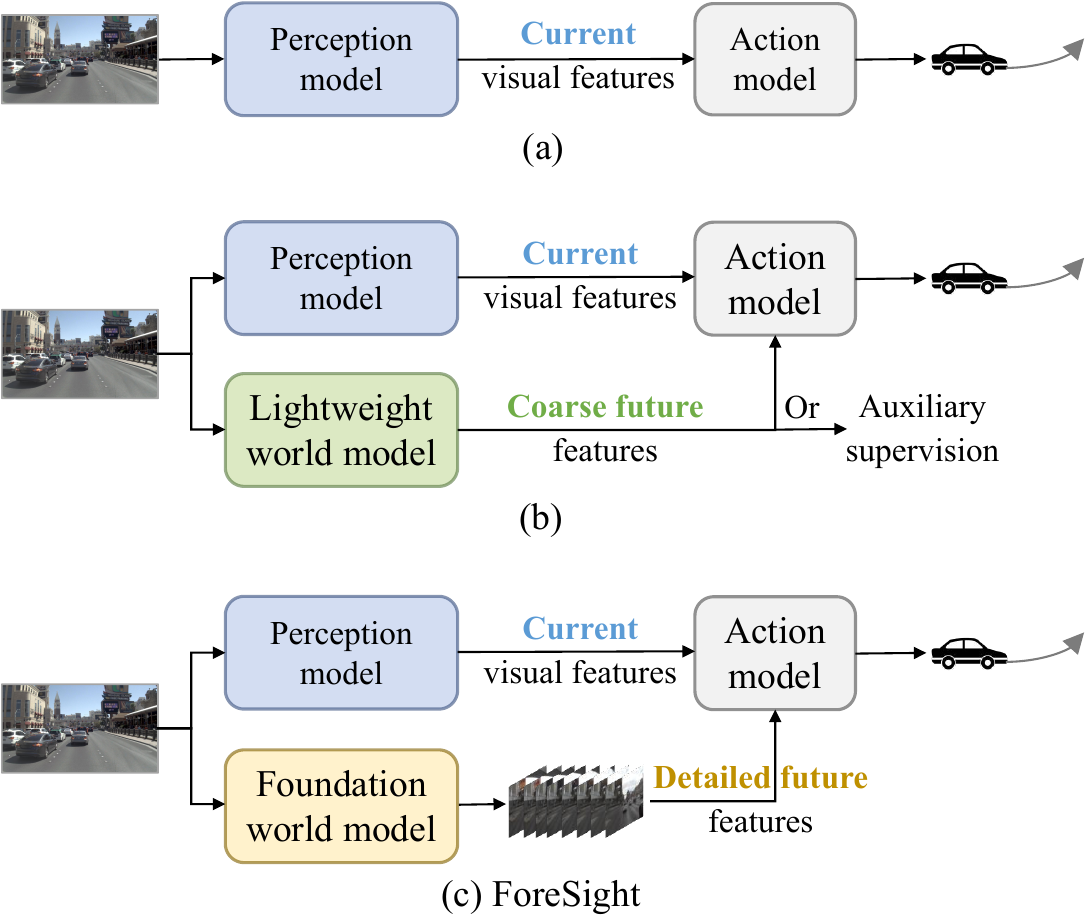}
\caption{
{\bf Paradigm comparison.}
(a) Reactive end-to-end planning based on current observations \cite{vad, UniAD,diffusiondrive}.
(b) A lightweight world model used as an auxiliary component for alignment or simplified prediction \cite{LAW, SSR,zhang2025seerdrive}.
(c) \textbf{\netName{}}: A foundation world model centric framework where future scene imagination drives action prediction.
}
\label{fig:cmp}
\end{figure}

As one of the most crucial tasks in autonomous driving, end-to-end planning has attracted widespread attention~\cite{nuscenes, navsim, Bench2Drive}, while the complexity of scene elements and their interactions significantly increases the difficulty of this task. Existing methods~\cite{TransFuser,vad,UniAD,sparsead,sparsedrive,goalflow,vadv2,drivetransformer,fu2025orion,diffusiondrive,diffusionplanner} follow a perception-to-planning paradigm, as shown in~\cref{fig:cmp} (a), predicting actions from historical and current observations. Despite achieving promising results, these methods are fundamentally reactive, as they decide based on what has happened and what is happening, rather than what will happen. Without explicitly imagining future scenes, they lack the foresight needed for proactive decision-making in dynamic traffic scenarios.

Foundation world models~\cite{liu2024sora, liao2025genie} have demonstrated strong capabilities in understanding and predicting dynamic evolution, inspiring their application to autonomous driving~\cite{gaia,russell2025gaia2,vista,zhang2025epona,wang2024driving,chen2025drivinggpt}. 
Hence, how to effectively integrate world models with planning frameworks is a worthwhile direction to explore.
However, existing approaches often treat world models as auxiliary components—either using them for representation alignment via reconstruction constraints or auxiliary supervision~\cite{SSR,LAW,zheng2025world4drive}, as shown in~\cref{fig:cmp} (b), or predicting simplified future features~\cite{wote, zhang2025seerdrive} without modeling full scene dynamics. Critically, these methods do not enable the planning module to genuinely leverage detailed future information as the primary basis for decision-making.

We argue that autonomous agents should imagine future scenes before deciding, just as human drivers mentally simulate ``what will happen next" before acting. We present \textbf{\netName{}}, a foundation world model centric planning framework that makes future scene imagination the primary driver of action prediction, as shown in~\cref{fig:cmp} (c). Unlike prior work, \netName{} integrates a pretrained world model directly as the visual encoder, generating detailed future scene visual representations that fundamentally inform the action decoder. This paradigm shift, from ``what should I do now?" to ``what will happen, and how should I respond?", enables genuinely \textit{anticipatory} rather than reactive planning. To complement future scene representations with current scene multi-modal and multi-view information, we employ a lightweight encoder for present observations. 
A state-based action decoder is designed for action prediction, which incorporates a WM-QFormer for aggregating and adapting future features and employs factorized attention to interact state-based trajectory queries with current and future scene features for planning.

Our {\bf contributions} are threefold: \textbf{(i)} We propose a foundation world model centric planning framework that makes future scene imagination the primary driver of action prediction, enabling \textit{anticipatory} rather than reactive autonomous driving. 
\textbf{(ii)} We realize this through \netName{}, which integrates a pretrained world model as the core visual encoder alongside a lightweight encoder for incorporating current scene observations and offering more scene context, as well as a specially designed action decoder for generating future trajectories.
\textbf{(iii)} Extensive experiments on NAVSIM and nuScenes benchmarks demonstrate that \netName{} achieves state-of-the-art performance.

\section{Related work}

\paragraph{End-to-end autonomous driving.}
As one of the most critical components in autonomous driving, end-to-end planning has witnessed remarkably rapid progress in recent years. Early approaches~\cite{TransFuser, stp3, UniAD, vad} follow the perception-to-planning paradigm, integrating multiple intermediate driving tasks into a unified framework. Building upon this design, subsequent works introduce sparse representations~\cite{sparsead, sparsedrive} and generative action modeling~\cite{diffusiondrive, goalflow} to enable more efficient scene understanding and more expressive motion patterns. To further enhance representational and generalization capabilities, recent studies~\cite{drivetransformer, vadv2,li2025ztrs,li2025generalized,hamdan2025eta,guo2025ipad,sima2025centaur,tang2025hip} have increasingly focused on aggregating scene information within Transformer-based architectures. Moreover, inspired by the success of multi-modal large language models (MLLMs), vision-language-action approaches~\cite{renz2025simlingo, fu2025orion, li2025recogdrive,li2025drivevla,zeng2025futuresightdrive,chi2025impromptu,zhou2025autovla,drivevlm,jiang2025alphadrive,cao2025fastdrivevla,chen2025solve} directly leverage the strengths of MLLMs to predict and plan driving actions.
Some other works explore the application of reinforcement learning~\cite{shang2025drivedpo,yan2025rlgf,chen2025rift,jaeger2025carl,li2025finetuning,jiang2025irl} in the field of autonomous driving.

\paragraph{World models in autonomous driving.}
Given the observations and states of the current environment, world models are capable of understanding and predicting future scene evolution.  
In the field of autonomous driving, generated content primarily focuses on realistic visual imagery~\cite{vista, gaia, zhang2025epona, ren2025cosmos, gao2025magicdrive, hu2024drivingworld}, road topology in the BEV space~\cite{sledge, gump, scenarioDreamer}, and scene-level occupancy representations~\cite{occworld,wei2024occllama,yang2025driving} that capture the spatial layout and dynamic evolution of surrounding agents and free space.
Regarding visual representations, methods such as~\cite{magicdrive, panacea, drivedreamer, Drivingintothefuture, vista,zhao2025pwm,yang2024drivearena,li2025uniscene,zhou2025hermes,yan2025drivingsphere,li2025omninwm} employ diffusion-based models to achieve controllable and realistic video generation. Specifically, Drive-WM~\cite{Drivingintothefuture} supports multi-view generation and explores its practical value in planning models, while Vista~\cite{vista} is trained on web-scale driving videos, endowing it with strong generative capability for diverse and long-horizon outputs. In contrast, the GAIA series~\cite{gaia, russell2025gaia2} and DrivingWorld~\cite{hu2024drivingworld} realize generative modeling in an auto-regressive manner, making them better suited for continuous long-term video generation. Moreover, Epona~\cite{zhang2025epona} integrates the strengths of both paradigms, enabling joint visual and motion modeling for comprehensive scene generation.

\paragraph{Planning with world modeling.}
Powerful world models can reveal the potential evolution of future scenes, providing a stronger foundation for motion prediction, as explored in robotics~\cite{hu2024vpp, liao2025genie}. The concepts of auxiliary supervision and collaborative optimization with world models have also been extended to autonomous driving planning frameworks. Specifically, auxiliary supervision-based approaches~\cite{SSR, LAW, zheng2025world4drive} leverage predicted action latents or trajectories as conditions to generate future visual representations via world models, which are then supervised using realistic representations. Meanwhile, collaborative optimization-based methods~\cite{wote, zhang2025seerdrive} employ BEV world models as trajectory selectors or to facilitate interaction with future information. However, these methods either rely on auxiliary supervision or resort to simplified features, making it difficult to fully exploit the powerful future representations provided by world models. In contrast, our approach directly takes the generated future representations as input, establishing more informative priors for action prediction.

\section{Methodology}
In this section, we introduce \netName{}, a foundation world model centric planning framework. We first introduce the task formulation and the overview of our pipeline in~\cref{sec::pre}. Then we describe two crucial representation sources in~\cref{sec::enc}, followed by our state-based interactive decoding for trajectory planning in~\cref{sec::dec}, and finally detail the two-phase model training strategy in~\cref{sec::train}.

\begin{figure*}[t!]
\centering
\includegraphics[width=0.98\textwidth]{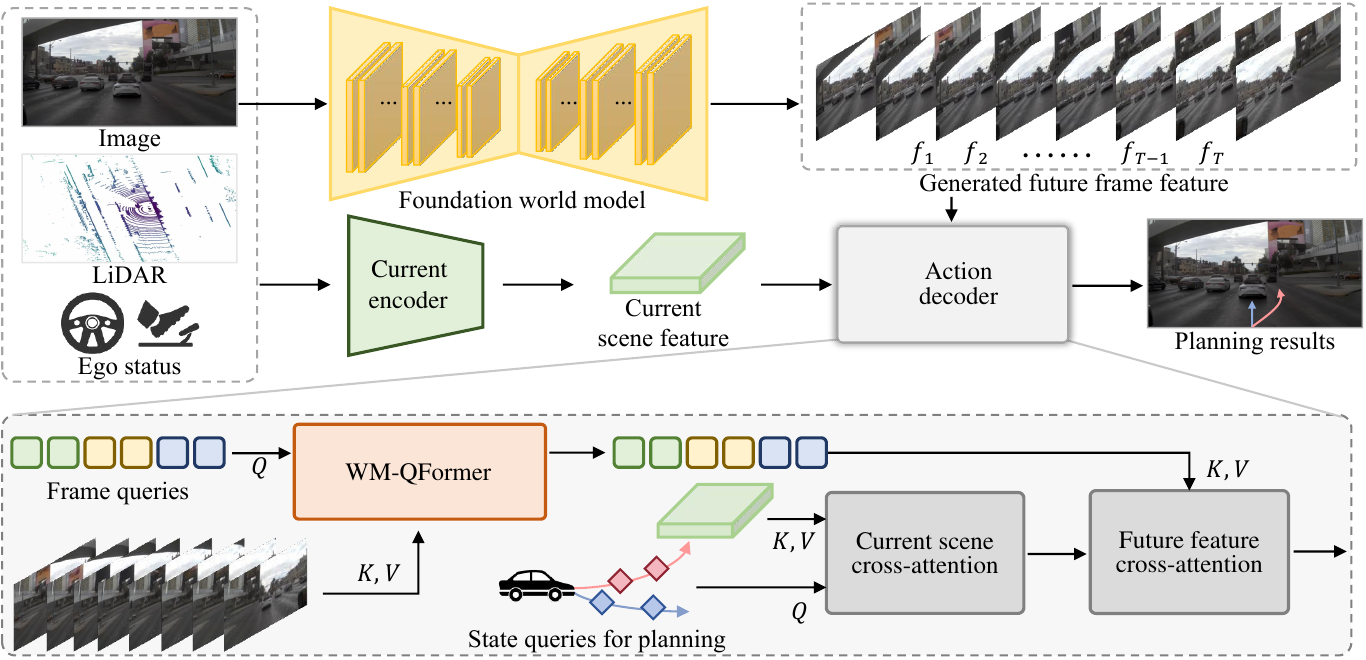}
\caption{
Overview of \textbf{\netName{}}. \netName{} introduces foundation world models into an end-to-end planning framework with using the current-frame features as an additional supplement. Besides, we design a WM-QFormer to compress future features with a set of frame queries and adapt them to the action head. To facilitate the interaction between action and visual presentations, we adopt state queries to explicitly represent time steps and factorized attention for feature interaction.
}
\label{fig:pipeline}
\end{figure*}

\subsection{Preliminary}
\label{sec::pre}

\paragraph{Task formulation.}
The end-to-end planning task in autonomous driving takes as input raw sensor data (e.g., multi-view images and LiDAR point clouds), captures the interactions among traffic elements, and finally predicts the future trajectories of the ego vehicle. To promote interpretability and reduce learning difficulty, multiple intermediate tasks are employed for auxiliary supervision, such as detection, map segmentation, and motion prediction. World modeling for autonomous driving aims to understand driving scenes and predict their dynamic evolution, which facilitates downstream applications such as real-world evaluation and simulation. In the context of our work, the world model is considered a powerful foresight generator that can provide rich future representations, serving the subsequent planning head to predict more accurate trajectories.

\paragraph{Pipeline overview.}
As depicted in~\cref{fig:pipeline}, the overall pipeline of \netName{} integrates a foundation world model into an end-to-end planning framework to enhance future reasoning and decision-making. Specifically, current-frame features are employed as an additional supplement to compensate for potential information gaps in the world model outputs. To align future representations with the action planning process, a WM-QFormer is introduced to compress the predicted future features through a set of frame queries and adapt them to the action head. Furthermore, state queries are utilized to explicitly encode temporal steps, enabling the model to capture the sequential nature of planning. To promote deeper interaction between visual and action representations, a factorized attention mechanism is applied, facilitating efficient cross-modal feature fusion and temporal reasoning throughout the pipeline.

\subsection{Input representation encoding}
\label{sec::enc}

The core vision encoder of \netName{} is the World Model (WM) encoder, which is directly inherited from existing foundation world models. For simplicity, we only consider diffusion-based world models in this work, such as~\cite{vista, zhang2025epona, Drivingintothefuture,ni2025maskgwm}. Specifically, the WM encoder takes as input the images of the current frame, conditioned on motion attributes (e.g., yaws and poses) or commands. During the denoising stage, a specific step $t_{\rm d}$ is selected to sample the latent features as future visual representations $F_{\rm wm} \in \mathbb{R}^{T_{\rm wm}\times C_{\rm wm} \times H \times W}$, where $T_{\rm wm}$ refers to the number of future frames, $C_{\rm wm}$ refers to the feature dimension, and $H$ and $W$ refer to the height and width of the feature maps. This process can be formulated as:
\begin{equation}
F_{\rm wm}= {\rm WM}^{(t_{\rm d})}(\mathcal{I}, F_{\rm cond}),
\end{equation}
where $\mathcal{I}$ and $F_{\rm cond}$ denote the raw input images and the condition latents, respectively.
Note, the sampling step $t_{\rm d}$ is adjustable to balance efficiency and performance. 

In addition, we also utilize a lightweight Transformer-based encoder~\cite{TransFuser,diffusiondrive,vad,SSR} for the current frame. In autonomous driving, accurately predicting future trajectories requires rich multi-view information, as relying solely on a forward-facing view can miss important cues from other directions. Most existing foundation world models primarily process front-view images, which may lead to incomplete or biased future representations. This lightweight encoder complements the multi-modal and cross-view information missing in the world model, providing precise and comprehensive features from the current frame, thereby further enhancing action decoding and improving the accuracy of predicted trajectories. Concretely, this module can be represented as:
\begin{equation}
F_{\rm cur}= {\rm Encoder}(\mathcal{I}, \mathcal{P}, \mathcal{E}),
\end{equation}
where $\mathcal{P}$ and $\mathcal{E}$ refer to the input LiDAR point cloud and ego status.

\subsection{State-based interactive decoding}
\label{sec::dec}

\paragraph{Time state queries representation for trajectory planning.}
The planning task in autonomous driving aims to generate multi-step future trajectories of the ego vehicle, which requires the model to reason over both spatial and temporal dimensions. Meanwhile, the foundation world model provides rich and temporally structured visual representations for the corresponding future time steps, offering valuable priors about scene evolution and interaction dynamics. To effectively align these temporally indexed future features with the planning process, we employ a set of learnable time state queries $Q_{\rm s}\in \mathbb{R}^{M\times T_{\rm f}\times C}$ following~\cite{bridgead}, where $M$ denotes the number of planning modes, $T_{\rm f}$ denotes the number of predicted future steps, and $C$ is the feature dimension. These queries explicitly encode temporal progression, enabling more precise and coherent interaction between future visual representations and the trajectory prediction process.

\paragraph{WM-QFormer for future feature aggregation.}
To aggregate scene information into queries, we introduce a dedicated WM-QFormer, specifically designed for processing the future world-model features $F_{\rm wm}$. The WM-QFormer is implemented as a spatial–temporal Transformer that jointly captures intra-frame spatial structure and inter-frame temporal dynamics. We use $N_{\rm wm}$ learnable queries for each frame to extract and compress the relevant information, producing compact representations $F'_{\rm wm} \in \mathbb{R}^{T_{\rm wm}\times N_{\rm wm} \times C}$.
This module is deliberately designed for world-model–based planning. The generated future frames typically contain abundant fine-grained textures and noise, which, if directly exposed to trajectory queries, may introduce interference into the planning process. Our WM-QFormer filters out these irrelevant details by selectively aggregating informative features from each frame, yielding a distilled representation that provides a reliable and noise-robust reference for the action head. This ensures that the planner benefits from meaningful future context while avoiding distraction from redundant or noisy visual cues.

\paragraph{Factorized attention for interacting trajectory queries with current and future features.}
Given the current features $F_{\rm cur}$, the compressed future features $F'_{\rm wm}$, and the state queries $Q_{\rm s}$, we adopt a factorized attention mechanism to enable effective interaction between the state queries and both current and future features through two separate cross-attention modules. Specifically, all state queries first attend to the current features via a standard cross-attention module, ensuring that each future time step comprehensively perceives the present scene. For the interaction with future features, we introduce additional time embeddings to encourage temporally adjacent steps to attend more strongly to each other, capturing the sequential dependencies in future trajectories. Moreover, considering that the future features and state queries may have different numbers of time steps, we employ sinusoidal positional embeddings to facilitate generalization across varying sequence lengths. The interaction process can be formulated as:
\begin{equation}
\begin{split}
Q_{\rm s} &= {\rm CrossAttn}(Q_{\rm s}, F_{\rm cur}), \\
Q_{\rm s} &= {\rm CrossAttn}(Q_{\rm s} + E_{\rm s}, F'_{\rm wm} + E_{\rm wm}), \\
\end{split}
\end{equation}
where $E_{\rm s}$ and $E_{\rm wm}$ denote the time embeddings of the state queries and future features, respectively. After the cross-attention interactions, a trajectory decoder is applied to produce the planned trajectories for the ego vehicle:
\begin{equation}
\mathcal{T} = {\rm TrajDecoder}(Q_{\rm s}),
\end{equation}
where $\mathcal{T}$ represents the decoded multi-step ego trajectories for planning. This factorized design allows the model to separately integrate present and predicted future information, thereby improving trajectory accuracy while maintaining temporal coherence.

\begin{table*} [t!]
\centering
\caption{
Performance comparison of planning on the NAVSIM \textit{navtest} split with closed-loop metrics.
The best and second-best results are highlighted in \textbf{bold} and \underline{underline}, respectively. We categorize the methods compared into three groups: planning models, world models, and planning with world models.
}
{\begin{tabular}{l|l|ccccc|c}
\toprule[1.5pt]
Type&Method & NC $\uparrow$ &DAC $\uparrow$ & TTC $\uparrow$& Comf. $\uparrow$ & EP $\uparrow$ & PDMS $\uparrow$ \\
\midrule
\multirow{9}{*}{Planning model}
&UniAD~\cite{UniAD}                                  & 97.8 & 91.9 & 92.9 & \textbf{100} & 78.8 & 83.4 \\
&PARA-Drive~\cite{paradrive}                         & 97.9 & 92.4 & 93.0 & \underline{99.8} & 79.3 & 84.0 \\
&TransFuser~\cite{TransFuser}                        & 97.7 & 92.8 & 92.8 & \textbf{100} & 79.2 & 84.0 \\
&DRAMA~\cite{drama}                                  & 98.0 & 93.1 & \underline{94.8} & \textbf{100} & 80.1 & 85.5 \\
&Hydra-MDP++~\cite{HydraMDPplus}                     & 97.6 & 96.0 & 93.1 & \textbf{100} & 80.4 & 86.6 \\
&DiffusionDrive~\cite{diffusiondrive}                & \underline{98.2} & 96.2 & 94.7 & \textbf{100} & 82.2 & 88.1 \\
&Hydra-NeXt~\cite{hydranext}                         & 98.1 & \underline{97.7} & 94.6 & \textbf{100} & 81.8 & 88.6 \\
&GoalFlow~\cite{goalflow} & \bf98.4 & \bf98.3 & 94.6 & \bf100 & \underline{85.0} & \underline{90.3} \\
&ReCogDrive~\cite{li2025recogdrive} & 97.9 & 97.3 & \bf94.9 & \bf100 & \textbf{87.3} & \bf90.8 \\

\midrule

\multirow{2}{*}{World model}&DrivingGPT~\cite{chen2025drivinggpt} & \bf98.9 & \underline{90.7} & \bf94.9 & \underline{95.6} & \underline{79.7} & \underline{82.4} \\
&Epona~\cite{zhang2025epona} & \underline{97.9} & \bf95.1 & \underline{93.8} & \bf99.9 & \bf80.4 & \bf86.2 \\

\midrule

\multirow{5}{*}{Planning with world model}&LAW~\cite{LAW}                                      & 96.4 & 95.4 & 88.7 & \underline{99.9} & 81.7 & 84.6 \\
&World4Drive~\cite{zheng2025world4drive} & 97.4 & 94.3 & 92.8 & \bf100 & 79.9 & 85.1\\
&WoTE~\cite{wote}                                    & \underline{98.5} & 96.8 & \bf94.9 & \underline{99.9} & 81.9 & 88.3 \\
&SeerDrive~\cite{zhang2025seerdrive}                      & 98.4 & \underline{97.0} & \bf 94.9 & \underline{99.9} & \underline{83.2} & \underline{88.9} \\
\rowcolor{gray!20}
& \bf\netName{}~(Ours)                              & \bf98.8 & \bf97.2 & \underline{94.8} & \bf100 & \bf 83.5 & \bf 89.3 \\

\bottomrule[1.5pt]
\end{tabular}}
\label{tab:navsim}
\end{table*}

\subsection{Model training}
\label{sec::train}

For efficient model training, we adopt a two-phase training strategy. In the first phase, we pretrain the original action model without incorporating world model features, allowing the model to effectively learn action-aware perceptual priors. After this action pretraining, we introduce the future visual representations from the world model and integrate the WM-QFormer modules into the pipeline. In the second phase, we perform post-training to jointly optimize all components. Notably, while the world model can undergo additional fine-tuning on the target dataset, it is kept fully frozen during our post-training stage, where the WM-QFormer is trained to adapt the future features to the action head. This two-phase strategy not only improves overall training efficiency but also alleviates instability in early training caused by the imbalance in representational capacity between the two feature sources.

We adopt the same training losses in both phases. The overall loss function $\mathcal{L}$ is defined as the weighted sum of the BEV segmentation loss $\mathcal{L}_{\rm bev}$ and the trajectory regression loss $\mathcal{L}_{\rm traj}$:
\begin{equation}
\mathcal{L} = \lambda_1\mathcal{L}_{\rm bev} + \lambda_2\mathcal{L}_{\rm traj}.
\end{equation}
where $\lambda_1$ and $\lambda_2$ denote the loss weights.

\section{Experiments}

\subsection{Datasets and metrics}
We perform experiments on the NAVSIM~\cite{navsim} and nuScenes~\cite{nuscenes} autonomous driving benchmarks. Specifically, NAVSIM is a subset of the nuPlan~\cite{nuplan} dataset, focusing on non-reactive simulation in complex scenarios with dynamic intention changes. The training/validation set and the testing set contain 1,192 and 136 scenarios, respectively, with a sampling frequency of 2 Hz for both camera and LiDAR data. Evaluation on NAVSIM is conducted using the PDM Score (PDMS), which is computed as a weighted sum of No At-Fault Collisions (NC), Drivable Area Compliance (DAC), Time-to-Collision (TTC), Comfort (Comf.), and Ego Progress (EP).
For the nuScenes dataset, it contains 1,000 scenes sampled at 2 Hz, and we adopt a 700/150 train/validation split following existing methods~\cite{UniAD, vad}. Evaluation on nuScenes is performed in an open-loop setting, and we use the metrics proposed in VAD~\cite{vad} for comparison.

\begin{table*} [t!] 
\centering
\caption{
Performance comparison of planning on the nuScenes \textit{validation} split. ResNet-50 is
used as the backbone for all the planning models, except for UniAD, which adopts ResNet-101.
}
{\begin{tabular}{l|l|cccc|cccc}
\toprule[1.5pt]
\multirow{2}{*}{Type} & \multirow{2}{*}{Method} &
\multicolumn{4}{c|}{L2 ($m$) $\downarrow$} & 
\multicolumn{4}{c}{Col. Rate (\%) $\downarrow$} \\
&& 1$s$ & 2$s$ & 3$s$ & Avg. & 1$s$ & 2$s$ & 3$s$ & Avg. \\
\midrule 

\multirow{8}{*}{Planning model}
&BEV-Planner~\cite{bevplanner}   & \underline{0.28} & \textbf{0.42} & \bf 0.68 & \textbf{0.46} & \underline{0.04} & 0.37 & 1.07 & 0.49 \\

&PARA-Drive~\cite{paradrive}     & \textbf{0.25} & \underline{0.46} & \underline{0.74} & \underline{0.48} & 0.14 & 0.23 & 0.39 & 0.25 \\
&VAD-Base~\cite{vad}             & 0.41 & 0.70 & 1.05 & 0.72 & 0.07 & 0.17 & 0.41 & 0.22 \\
&GenAD~\cite{genad}              & \underline{0.28} & 0.49 & 0.78 & 0.52 & 0.08 & 0.14 & 0.34 & 0.19 \\
&UniAD~\cite{UniAD}              & 0.44 & 0.67 & 0.96 & 0.69 & \underline{0.04} & \underline{0.08} & 0.23 & 0.12 \\
&BridgeAD~\cite{bridgead}        & 0.29 & 0.57 & 0.92 & 0.59 & \textbf{0.01} & \bf 0.05 & \underline{0.22} & \underline{0.09} \\
&MomAD~\cite{momad}              & 0.31 & 0.57 & 0.91 & 0.60 & \textbf{0.01} & \bf 0.05 & \underline{0.22} & \underline{0.09} \\
&SparseDrive~\cite{sparsedrive}  & 0.29 & 0.58 & 0.96 & 0.61 & \textbf{0.01} & \bf 0.05 & \bf0.18 & \bf 0.08 \\
\midrule
\multirow{3}{*}{Planning with world model}&LAW~\cite{LAW}                  & \underline{0.26} & 0.57 & 1.01 & \underline{0.61} & 0.14 & \underline{0.21} & 0.54 & 0.30 \\
&World4Drive~\cite{zheng2025world4drive} & \bf0.23 & \bf0.47 & \bf0.81 & \bf0.50 & \textbf{0.02} & \bf0.12 & \bf0.33 & \bf0.16 \\

\rowcolor{gray!20}
&\bf\netName~(Ours)                 & 0.36 & \underline{0.55} & \underline{0.93} & 0.62 & \underline{0.04} & \bf0.12 & \underline{0.37} & \underline{0.18} \\

\bottomrule[1.5pt]
\end{tabular}}


\label{tab:nusc}
\end{table*}

\subsection{Implementation details}

\paragraph{Model settings.} 
For the NAVSIM benchmark, we utilize both images and LiDAR as raw input, while only images are used for the nuScenes dataset. Specifically, 3 camera views are used for NAVSIM, whereas 6 views are used for nuScenes. The resolution of input images is set to 1024×256 for NAVSIM following previous methods~\cite{TransFuser, diffusiondrive}, while it is downscaled to 640×360 for the nuScenes dataset.

Regarding our choice of world models, we mainly employ Epona~\cite{zhang2025epona} for NAVSIM and nuScenes, which enables direct adaptation without requiring any fine-tuning or minor adjustments to the output frequency. 
We generate future frames from the world model with the same number of steps as used in planning — 8 future frames for NAVSIM and 6 for nuScenes — both conditioned only on the current frame.
To further demonstrate the wide adaptability of our model, we also use Vista~\cite{vista} as the world model on nuScenes, with the results presented in the supplementary materials.

As for the output trajectories, planning needs to consider the multimodality of future trajectories. For the NAVSIM dataset, a 4-second planning window with 8 future steps is used, and we adopt a common setting of 20 trajectory modes for decoding. For the nuScenes dataset, the planning window is set to 3 seconds with 6 future steps, and 6 modes are utilized for decoding.

\paragraph{Training settings.} 
We train our model using 8 NVIDIA H100 GPUs. For NAVSIM, the batch size is 8, with 80 epochs for pretraining the action model without using the world model, followed by 20 epochs for post-training the final model combining the world model and the action model. For nuScenes, the batch size is 1, with 12 epochs for pretraining and 6 epochs for post-training. The learning rate is set to $1 \times 10^{-4}$ for both NAVSIM and nuScenes, both of which are optimized using AdamW~\cite{adamw}.

\begin{table*} [ht!]
\centering
\caption{Ablation study on the key components of our model. ``w. WM" indicates that the world model is incorporated to facilitate the planning process.}

{\begin{tabular}{c|cccc|cccccc}
\toprule[1.5pt]
\multirow{2}{*}{ID} & \multirow{2}{*}{w. WM} & \multirow{2}{*}{WM-QFormer}& State & Factorized& \multirow{2}{*}{NC $\uparrow$} &\multirow{2}{*}{DAC $\uparrow$} & \multirow{2}{*}{TTC $\uparrow$}& \multirow{2}{*}{Comf. $\uparrow$} & \multirow{2}{*}{EP $\uparrow$} & \multirow{2}{*}{PDMS $\uparrow$}  \\
&& & queries &attetntion & & & &\\
\midrule
1 &&               & && 97.8 & 95.6 & 93.4 & \bf100 & 81.6 & \cellcolor{gray!20} 86.8 \\
2 & \checkmark &    & && 97.8 & 95.9 & 93.3 & \bf100 & 82.1 & \cellcolor{gray!20} 87.1 \\
3 &\checkmark &  \checkmark  & && 98.6 & 96.2 & 95.0 & \bf100 & 81.3 & \cellcolor{gray!20} 87.9 \\
4 &\checkmark &  \checkmark  &\checkmark & & 98.6 & 96.8 & 95.0 & \bf100 & 82.0 & \cellcolor{gray!20} 88.5 \\
5 &\checkmark &  \checkmark  & &\checkmark& 98.4 & 96.5 & \bf95.3 & \bf100 & 81.6 & \cellcolor{gray!20} 88.2 \\

\rowcolor{gray!20}
6 &\checkmark & \checkmark &\checkmark&\checkmark& \bf98.8 & \bf97.2 & 94.8 & \bf100 & \bf83.5 & \bf89.3 \\

\bottomrule[1.5pt]
\end{tabular}}

\label{tab:abl_main}
\end{table*}

\begin{table} [ht!]
\centering
\caption{
Ablation study on the use of different types of world models. ``Simple WM'' refers to lightweight or simplified world models, such as~\cite{wote, zhang2025seerdrive}, while ``Found. WM'' denotes the foundation world models employed in our method.}

{\begin{tabular}{c|cccc}
\toprule[1.5pt]
 & DAC $\uparrow$ & TTC $\uparrow$ & EP $\uparrow$ &  PDMS $\uparrow$  \\
\midrule

w/o WM & 95.6 & 93.4 & 81.6 & \cellcolor{gray!20} 86.8 \\
Simple WM & 96.3 & 93.4 & 82.2 & \cellcolor{gray!20} 87.5 \\

\rowcolor{gray!20}
Found. WM &  \bf97.2 & \bf94.8 & \bf83.5 & \bf89.3  \\

\bottomrule[1.5pt]
\end{tabular}}

\label{tab:abl_wm}
\end{table}

\begin{table}[ht!]
\centering
\caption{Ablation study on the number of steps in the denoising procedure.}

{\begin{tabular}{c|cccc}
\toprule[1.5pt]
Steps & DAC $\uparrow$ & TTC $\uparrow$ & EP $\uparrow$ &  PDMS $\uparrow$ \\
\midrule

25 & 96.4 & 94.8 & 81.3 & \cellcolor{gray!20} 88.0 \\
50 & 96.6 & \bf95.2 & 81.5 & \cellcolor{gray!20} 88.3 \\
75 & \bf97.3 & 94.7 & \bf83.5 & \cellcolor{gray!20} 89.2 \\
\rowcolor{gray!20}
100 & 97.2 & 94.8 & \bf83.5 & \cellcolor{gray!20} \bf89.3 \\

\bottomrule[1.5pt]
\end{tabular}}

\label{tab:abl_dnstep}
\end{table}

\subsection{Comparison with state of the art}
As shown in~\cref{tab:navsim}, we compare our \netName{} with several state-of-the-art methods on the NAVSIM \textit{navtest} split. To ensure a fair comparison, existing methods are divided into three groups: planning models, world models, and planning with world models. Our approach achieves the highest PDM Score of 89.3, obtaining the best results within the planning-with-world-models group. It significantly outperforms other methods, benefiting from the scene understanding and future representations provided by foundation world models.
Compared with the world model group, where the planning branch primarily serves as an auxiliary component to enhance the generation process, our \netName{} significantly surpasses DrivingGPT~\cite{chen2025drivinggpt} and Epona~\cite{zhang2025epona}, demonstrating the effectiveness of our specially designed action decoder.
Compared with planning model methods, except for the powerful VLA-based approach ReCogDrive~\cite{li2025recogdrive} trained with reinforcement learning and GoalFlow equipped with a V2-99 backbone, our method outperforms all other planning methods.

Furthermore, we evaluate our method on the open-loop nuScenes dataset, as shown in~\cref{tab:nusc}. Although the scenarios and evaluation protocols in nuScenes are relatively simple and the metrics are not entirely comprehensive~\cite{bevplanner}, the dataset still provides a suitable environment for validating our approach. Compared with recent state-of-the-art methods, including powerful planning models such as PARA-Drive~\cite{paradrive}, BridgeAD~\cite{bridgead}, SparseDrive~\cite{sparsedrive}, and MomAD~\cite{momad}, our method demonstrates competitive performance.

\subsection{Ablation study}

\paragraph{Effects of components.}
We first conduct an ablation study on all key components of \netName{}, as shown in~\cref{tab:abl_main}. The first row represents our baseline, which uses only the current visual encoder and a simple action decoder, achieving a PDMS of 86.8. In the second row, we integrate the world model into the framework and interact future frame features using a vanilla attention mechanism. This yields a slight improvement over the baseline, indicating that future features can benefit the planning process, but not in a straightforward manner.
In the third row, we design the WM-QFormer to aggregate the generated future frame features, resulting in a significant improvement in the PDMS to 87.9. This demonstrates that the WM-QFormer effectively aggregates relevant information for planning while mitigating interfering factors.
The fourth and fifth rows illustrate the impact of the state queries and the factorized attention module. Each component individually contributes to performance gains, and their combination leads to an even more substantial improvement. The state queries and factorized attention work together to provide precise representations of future trajectories and enable accurate interactions with the generated future frame features. All components are integrated in the last row, forming the complete model and achieving optimal performance.

\paragraph{Effects of foundation world model.}
To validate the effectiveness of large-scale pretrained foundation world models, we also explore replacing them with lightweight future predictors such as~\cite{wote, SSR, zhang2025seerdrive}. The experimental results in~\cref{tab:abl_wm} show that foundation world models significantly outperform these simplified predictors. Although the lightweight predictors can learn future representations to some extent after training, they struggle to generalize to complex or edge-case scenarios, limiting their impact on downstream planning. In contrast, foundation world models exhibit strong generalization and generative capabilities, effectively mitigating this issue. As a result, they can substantially assist the action head in understanding future scenarios and guiding decision-making.

\paragraph{Effects of the number of denoising steps.}
We investigate the impact of the number of denoising steps, as shown in~\cref{tab:abl_dnstep}. In the early stages of denoising, future visual representations evolve from coarse to fine, which significantly benefits action planning. In later steps, the representations have already captured the essential traffic elements and undergo only minor refinements. Consequently, additional steps provide marginal improvements while increasing inference time. While the final model uses 100 denoising steps to achieve the highest performance, 75 steps suffice to strike a favorable balance between efficiency and accuracy, yielding the best trade-off.

\subsection{Qualitative results}
\begin{figure*}[t!]
\centering
\includegraphics[width=0.98\textwidth]{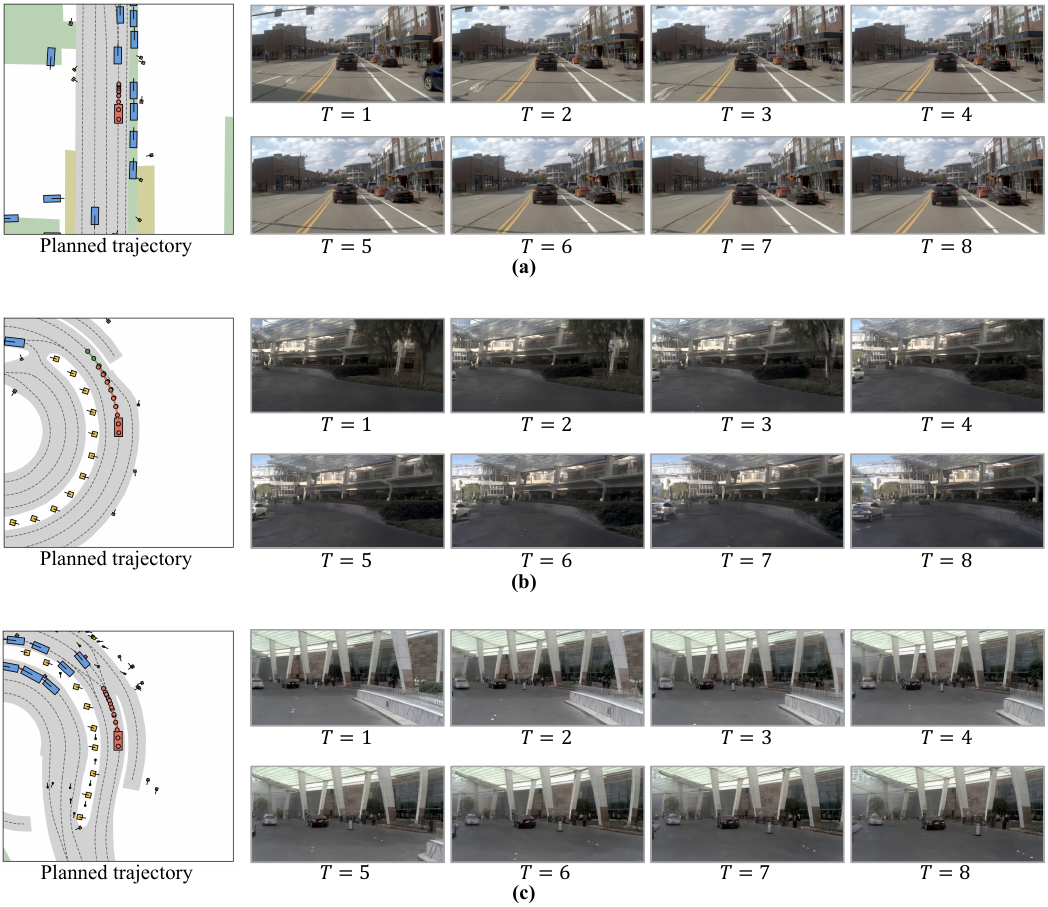}
\caption{
Visualization of \textbf{\netName{}} on the NAVSIM~\cite{navsim} dataset. The left panel shows the planned trajectories in the BEV view, while the right panel presents the generated future video over the next 8 time steps. The ground-truth trajectory is depicted in \textcolor{ForestGreen}{green}, and the final planned trajectory is highlighted in \textcolor{orange}{orange}.
}
\label{fig:visual}
\end{figure*}

As shown in~\cref{fig:visual}, we visualize the planned trajectories alongside the corresponding 8 future frames generated by the world models. Interaction scenarios are illustrated in (a) and (c), where the interactions between traffic elements in the future scenes are generated appropriately. This assists the planning model in better understanding the scene and modeling relationships, ultimately yielding accurate trajectories for the ego vehicle. Moreover, given the importance of generated future visual representations for action prediction, we also visualize complex scenarios, such as turning behaviors, in (b) and (c). It can be observed that the foundation world model effectively handles these situations, generating future frames that guide the final trajectories. Additional visualizations and failure cases are provided in the supplementary materials.

\section{Conclusion}
We present \netName{}, a foundation world model centric framework for end-to-end autonomous driving planning that shifts the paradigm from reactive to anticipatory decision-making. By explicitly imagining plausible future scenes, \netName{} enables autonomous agents to grasp upcoming dynamics and respond accordingly, mirroring human driving behavior. The framework leverages a pretrained world model to generate detailed future scene representations, complemented by a lightweight encoder for current observations. These features are integrated through a state-based decoder equipped with a WM-QFormer for future feature aggregation and factorized attention to enable effective interaction between state-based trajectory queries and both current and future scene representations. Extensive evaluations on the NAVSIM and nuScenes benchmarks show that planning with anticipated future contexts significantly outperforms state-of-the-art methods, validating the benefits of foresight-driven planning. \netName{} demonstrates the promise of coupling foundation world models with planning modules for more intelligent and anticipatory autonomous driving systems.

\noindent\textbf{Limitations and future work.} 
\netName{} relies on two key components: the world model and the action head. The world model offers strong scene understanding and future representations but remains computationally expensive, limiting practicality. For the action head, recent end-to-end and VLA-based approaches have leveraged reinforcement learning to boost performance. Incorporating similar strategies into the world–action paradigm may further enhance action generation and overall planning.

\section*{Acknowledgements}
This work was supported in part by New Generation Artificial Intelligence-National Science and Technology Major Project (2025ZD0123004), Ningbo grant (2025Z038) and National Natural Science Foundation of China (Grant No. 62376060).



{
    \small
    \bibliographystyle{ieeenat_fullname}
    \bibliography{main}

@String(CVPR= {IEEE Conf. Comput. Vis. Pattern Recog.})

@String(ICCV= {Int. Conf. Comput. Vis.})

@String(ECCV= {Eur. Conf. Comput. Vis.})

@String(NIPS= {Adv. Neural Inform. Process. Syst.})

@String(ICLR = {Int. Conf. Learn. Represent.})

@String(AAAI = {AAAI})

@String(CVPR  = {CVPR})

@String(ICML  = {ICML})

@String(ICCV  = {ICCV})

@String(ECCV  = {ECCV})

@String(NIPS  = {NeurIPS})

@String(ICLR  = {ICLR})

@String(CoRL = {CoRL})

@String(ICRA = {ICRA})

@String(NeurIPS  = {NeurIPS})

@inproceedings{UniAD,
  title={Planning-oriented autonomous driving},
  author={Hu, Yihan and Yang, Jiazhi and Chen, Li and Li, Keyu and Sima, Chonghao and Zhu, Xizhou and Chai, Siqi and Du, Senyao and Lin, Tianwei and Wang, Wenhai and others},
  booktitle=CVPR,
  year={2023}
}

@inproceedings{vad,
  title={Vad: Vectorized scene representation for efficient autonomous driving},
  author={Jiang, Bo and Chen, Shaoyu and Xu, Qing and Liao, Bencheng and Chen, Jiajie and Zhou, Helong and Zhang, Qian and Liu, Wenyu and Huang, Chang and Wang, Xinggang},
  booktitle=ICCV,
  year={2023}
}

@article{sparsead,
  title={SparseAD: Sparse Query-Centric Paradigm for Efficient End-to-End Autonomous Driving},
  author={Zhang, Diankun and Wang, Guoan and Zhu, Runwen and Zhao, Jianbo and Chen, Xiwu and Zhang, Siyu and Gong, Jiahao and Zhou, Qibin and Zhang, Wenyuan and Wang, Ningzi and others},
  journal={arXiv preprint},
  year={2024}
}

@inproceedings{LAW,
  title={Enhancing End-to-End Autonomous Driving with Latent World Model},
  author={Li, Yingyan and Fan, Lue and He, Jiawei and Wang, Yuqi and Chen, Yuntao and Zhang, Zhaoxiang and Tan, Tieniu},
  booktitle=ICLR,
  year={2025}
}

@article{vadv2,
  title={Vadv2: End-to-end vectorized autonomous driving via probabilistic planning},
  author={Chen, Shaoyu and Jiang, Bo and Gao, Hao and Liao, Bencheng and Xu, Qing and Zhang, Qian and Huang, Chang and Liu, Wenyu and Wang, Xinggang},
  journal={arXiv preprint},
  year={2024}
}

@inproceedings{Drivingintothefuture,
  title={Driving into the future: Multiview visual forecasting and planning with world model for autonomous driving},
  author={Wang, Yuqi and He, Jiawei and Fan, Lue and Li, Hongxin and Chen, Yuntao and Zhang, Zhaoxiang},
  booktitle=CVPR,
  year={2024}
}

@inproceedings{occworld,
    title={OccWorld: Learning a 3D Occupancy World Model for Autonomous Driving},
    author={Zheng, Wenzhao and Chen, Weiliang and Huang, Yuanhui and Zhang, Borui and Duan, Yueqi and Lu, Jiwen },
    booktitle=ECCV,
    year={2024}
}

@inproceedings{drivevlm,
  title={Drivevlm: The convergence of autonomous driving and large vision-language models},
  author={Tian, Xiaoyu and Gu, Junru and Li, Bailin and Liu, Yicheng and Hu, Chenxu and Wang, Yang and Zhan, Kun and Jia, Peng and Lang, Xianpeng and Zhao, Hang},
  booktitle=CoRL,
  year={2024}
}

@inproceedings{genad,
    title={GenAD: Generative End-to-End Autonomous Driving},
    author={Zheng, Wenzhao and Song, Ruiqi and Guo, Xianda and Zhang, Chenming and Chen, Long},
    booktitle=ECCV,
    year={2024}
}

@inproceedings{sparsedrive,
  title={SparseDrive: End-to-End Autonomous Driving via Sparse Scene Representation},
  author={Sun, Wenchao and Lin, Xuewu and Shi, Yining and Zhang, Chuang and Wu, Haoran and Zheng, Sifa},
  booktitle=ICRA,
  year={2025}
}

@inproceedings{stp3,
  title={St-p3: End-to-end vision-based autonomous driving via spatial-temporal feature learning},
  author={Hu, Shengchao and Chen, Li and Wu, Penghao and Li, Hongyang and Yan, Junchi and Tao, Dacheng},
  booktitle=ECCV,
  year={2022},
}

@inproceedings{nuscenes,
  title={nuscenes: A multimodal dataset for autonomous driving},
  author={Caesar, Holger and Bankiti, Varun and Lang, Alex H and Vora, Sourabh and Liong, Venice Erin and Xu, Qiang and Krishnan, Anush and Pan, Yu and Baldan, Giancarlo and Beijbom, Oscar},
  booktitle=CVPR,
  year={2020}
}

@inproceedings{bevplanner,
  title={Is ego status all you need for open-loop end-to-end autonomous driving?},
  author={Li, Zhiqi and Yu, Zhiding and Lan, Shiyi and Li, Jiahan and Kautz, Jan and Lu, Tong and Alvarez, Jose M},
  booktitle=CVPR,
  year={2024}
}

@inproceedings{adamw,
  title={Decoupled weight decay regularization},
  author={Loshchilov, I},
  booktitle=ICLR,
  year={2019}
}

@article{nuplan,
  title={nuplan: A closed-loop ml-based planning benchmark for autonomous vehicles},
  author={Caesar, Holger and Kabzan, Juraj and Tan, Kok Seang and Fong, Whye Kit and Wolff, Eric and Lang, Alex and Fletcher, Luke and Beijbom, Oscar and Omari, Sammy},
  journal={arXiv preprint},
  year={2021}
}

@inproceedings{TransFuser,
  title={Multi-modal fusion transformer for end-to-end autonomous driving},
  author={Prakash, Aditya and Chitta, Kashyap and Geiger, Andreas},
  booktitle=CVPR,
  year={2021}
}

@inproceedings{Bench2Drive,
  title={Bench2Drive: Towards Multi-Ability Benchmarking of Closed-Loop End-To-End Autonomous Driving},
  author={Xiaosong Jia and Zhenjie Yang and Qifeng Li and Zhiyuan Zhang and Junchi Yan},
  booktitle=NeurIPS,
  year={2024}
}

@inproceedings{diffusiondrive,
  title={DiffusionDrive: Truncated Diffusion Model for End-to-End Autonomous Driving},
  author={Bencheng Liao and Shaoyu Chen and Haoran Yin and Bo Jiang and Cheng Wang and Sixu Yan and Xinbang Zhang and Xiangyu Li and Ying Zhang and Qian Zhang and Xinggang Wang},
   booktitle=CVPR,
   year={2025},
}

@inproceedings{diffusionplanner,
    title={Diffusion-Based Planning for Autonomous Driving with Flexible Guidance},
    author={Yinan Zheng and Ruiming Liang and Kexin ZHENG and Jinliang Zheng and Liyuan Mao and Jianxiong Li and Weihao Gu and Rui Ai and Shengbo Eben Li and Xianyuan Zhan and Jingjing Liu},
    booktitle=ICLR,
    year={2025},
}

@inproceedings{navsim,
  title={Navsim: Data-driven non-reactive autonomous vehicle simulation and benchmarking},
  author={Dauner, Daniel and Hallgarten, Marcel and Li, Tianyu and Weng, Xinshuo and Huang, Zhiyu and Yang, Zetong and Li, Hongyang and Gilitschenski, Igor and Ivanovic, Boris and Pavone, Marco and others},
  booktitle=NIPS,
  year={2024},
}

@inproceedings{SSR,
  title={Navigation-Guided Sparse Scene Representation for End-to-End Autonomous Driving},
  author={Peidong Li and Dixiao Cui},
  booktitle=ICLR,
  year={2025}
}

@inproceedings{drivetransformer,
  title={DriveTransformer: Unified Transformer for Scalable End-to-End Autonomous Driving},
  author={Xiaosong Jia and Junqi You and Zhiyuan Zhang and Junchi Yan},
  booktitle=ICLR,
  year={2025}
}

@inproceedings{goalflow,
  title={GoalFlow: Goal-Driven Flow Matching for Multimodal Trajectories Generation in End-to-End Autonomous Driving},
  author={Xing, Zebin and Zhang, Xingyu and Hu, Yang and Jiang, Bo and He, Tong and Zhang, Qian and Long, Xiaoxiao and Yin, Wei},
  booktitle=CVPR,
  year={2025},
}

@inproceedings{magicdrive,
  title={MagicDrive: Street View Generation with Diverse 3D Geometry Control},
  author={Gao, Ruiyuan and Chen, Kai and Xie, Enze and Lanqing, HONG and Li, Zhenguo and Yeung, Dit-Yan and Xu, Qiang},
  booktitle=ICLR,
  year={2024}
}

@inproceedings{panacea,
  title={Panacea: Panoramic and controllable video generation for autonomous driving},
  author={Wen, Yuqing and Zhao, Yucheng and Liu, Yingfei and Jia, Fan and Wang, Yanhui and Luo, Chong and Zhang, Chi and Wang, Tiancai and Sun, Xiaoyan and Zhang, Xiangyu},
  booktitle=CVPR,
  year={2024}
}

@inproceedings{drivedreamer,
  title={DriveDreamer: Towards Real-World-Drive World Models for Autonomous Driving},
  author={Wang, Xiaofeng and Zhu, Zheng and Huang, Guan and Chen, Xinze and Zhu, Jiagang and Lu, Jiwen},
  booktitle=ECCV,
  year={2024},
}

@inproceedings{vista,
  title={Vista: A Generalizable Driving World Model with High Fidelity and Versatile Controllability},
  author={Gao, Shenyuan and Yang, Jiazhi and Chen, Li and Chitta, Kashyap and Qiu, Yihang and Geiger, Andreas and Zhang, Jun and Li, Hongyang},
  booktitle=NIPS,
  year={2024},
}

@article{gaia,
  title={Gaia-1: A generative world model for autonomous driving},
  author={Hu, Anthony and Russell, Lloyd and Yeo, Hudson and Murez, Zak and Fedoseev, George and Kendall, Alex and Shotton, Jamie and Corrado, Gianluca},
  journal={arXiv preprint},
  year={2023}
}

@inproceedings{sledge,
  title={Sledge: Synthesizing driving environments with generative models and rule-based traffic},
  author={Chitta, Kashyap and Dauner, Daniel and Geiger, Andreas},
  booktitle=ECCV,
  year={2024},
}

@inproceedings{scenarioDreamer,
  title={Scenario Dreamer: Vectorized Latent Diffusion for Generating Driving Simulation Environments},
  author={Rowe, Luke and Girgis, Roger and Gosselin, Anthony and Paull, Liam and Pal, Christopher and Heide, Felix},
  booktitle=CVPR,
  year={2025},
}

@inproceedings{gump,
  title={Solving motion planning tasks with a scalable generative model},
  author={Hu, Yihan and Chai, Siqi and Yang, Zhening and Qian, Jingyu and Li, Kun and Shao, Wenxin and Zhang, Haichao and Xu, Wei and Liu, Qiang},
  booktitle=ECCV,
  year={2024},
}

@article{wote,
  title={End-to-End Driving with Online Trajectory Evaluation via BEV World Model},
  author={Li, Yingyan and Wang, Yuqi and Liu, Yang and He, Jiawei and Fan, Lue and Zhang, Zhaoxiang},
  journal={arXiv preprint},
  year={2025}
}

@article{HydraMDPplus,
  title={Hydra-MDP++: Advancing End-to-End Driving via Expert-Guided Hydra-Distillation},
  author={Li, Kailin and Li, Zhenxin and Lan, Shiyi and Xie, Yuan and Zhang, Zhizhong and Liu, Jiayi and Wu, Zuxuan and Yu, Zhiding and Alvarez, Jose M},
  journal={arXiv preprint},
  year={2025}
}

@article{drama,
  title={Drama: An efficient end-to-end motion planner for autonomous driving with mamba},
  author={Yuan, Chengran and Zhang, Zhanqi and Sun, Jiawei and Sun, Shuo and Huang, Zefan and Lee, Christina Dao Wen and Li, Dongen and Han, Yuhang and Wong, Anthony and Tee, Keng Peng and others},
  journal={arXiv preprint},
  year={2024}
}

@article{hydranext,
  title={Hydra-next: Robust closed-loop driving with open-loop training},
  author={Li, Zhenxin and Wang, Shihao and Lan, Shiyi and Yu, Zhiding and Wu, Zuxuan and Alvarez, Jose M},
  journal={arXiv preprint},
  year={2025}
}

@inproceedings{bridgead,
 title={Bridging Past and Future: End-to-End Autonomous Driving with Historical Prediction and Planning},
 author={Zhang, Bozhou and Song, Nan and Jin, Xin and Zhang, Li},
 booktitle={CVPR},
 year={2025},
}

@inproceedings{momad,
  title={Don't Shake the Wheel: Momentum-Aware Planning in End-to-End Autonomous Driving},
  author={Song, Ziying and Jia, Caiyan and Liu, Lin and Pan, Hongyu and Zhang, Yongchang and Wang, Junming and Zhang, Xingyu and Xu, Shaoqing and Yang, Lei and Luo, Yadan},
  booktitle={CVPR},
  year={2025}
}

@inproceedings{paradrive,
  title={PARA-Drive: Parallelized Architecture for Real-time Autonomous Driving},
  author={Weng, Xinshuo and Ivanovic, Boris and Wang, Yan and Wang, Yue and Pavone, Marco},
  booktitle=CVPR,
  year={2024}
}

@inproceedings{renz2025simlingo,
  title={Simlingo: Vision-only closed-loop autonomous driving with language-action alignment},
  author={Renz, Katrin and Chen, Long and Arani, Elahe and Sinavski, Oleg},
  booktitle=cvpr,
  year={2025}
}

@article{li2025recogdrive,
  title={Recogdrive: A reinforced cognitive framework for end-to-end autonomous driving},
  author={Li, Yongkang and Xiong, Kaixin and Guo, Xiangyu and Li, Fang and Yan, Sixu and Xu, Gangwei and Zhou, Lijun and Chen, Long and Sun, Haiyang and Wang, Bing and others},
  journal={arXiv preprint},
  year={2025}
}

@article{fu2025orion,
  title={Orion: A holistic end-to-end autonomous driving framework by vision-language instructed action generation},
  author={Fu, Haoyu and Zhang, Diankun and Zhao, Zongchuang and Cui, Jianfeng and Liang, Dingkang and Zhang, Chong and Zhang, Dingyuan and Xie, Hongwei and Wang, Bing and Bai, Xiang},
  journal=ICCV,
  year={2025}
}

@article{hu2024drivingworld,
  title={DrivingWorld: Constructing world model for autonomous driving via video GPT},
  author={Hu, Xiaotao and Yin, Wei and Jia, Mingkai and Deng, Junyuan and Guo, Xiaoyang and Zhang, Qian and Long, Xiaoxiao and Tan, Ping},
  journal={arXiv preprint},
  year={2024}
}

@article{russell2025gaia2,
  title={Gaia-2: A controllable multi-view generative world model for autonomous driving},
  author={Russell, Lloyd and Hu, Anthony and Bertoni, Lorenzo and Fedoseev, George and Shotton, Jamie and Arani, Elahe and Corrado, Gianluca},
  journal={arXiv preprint},
  year={2025}
}

@inproceedings{zhang2025epona,
  title={Epona: Autoregressive Diffusion World Model for Autonomous Driving},
  author={Zhang, Kaiwen and Tang, Zhenyu and Hu, Xiaotao and Pan, Xingang and Guo, Xiaoyang and Liu, Yuan and Huang, Jingwei and Yuan, Li and Zhang, Qian and Long, Xiao-Xiao and others},
  booktitle=ICCV,
  year={2025}
}

@article{hu2024vpp,
  title={Video prediction policy: A generalist robot policy with predictive visual representations},
  author={Hu, Yucheng and Guo, Yanjiang and Wang, Pengchao and Chen, Xiaoyu and Wang, Yen-Jen and Zhang, Jianke and Sreenath, Koushil and Lu, Chaochao and Chen, Jianyu},
  journal=ICML,
  year={2025}
}

@article{liao2025genie,
  title={Genie envisioner: A unified world foundation platform for robotic manipulation},
  author={Liao, Yue and Zhou, Pengfei and Huang, Siyuan and Yang, Donglin and Chen, Shengcong and Jiang, Yuxin and Hu, Yue and Cai, Jingbin and Liu, Si and Luo, Jianlan and others},
  journal={arXiv preprint},
  year={2025}
}

@inproceedings{zheng2025world4drive,
  title={World4Drive: End-to-end autonomous driving via intention-aware physical latent world model},
  author={Zheng, Yupeng and Yang, Pengxuan and Xing, Zebin and Zhang, Qichao and Zheng, Yuhang and Gao, Yinfeng and Li, Pengfei and Zhang, Teng and Xia, Zhongpu and Jia, Peng and others},
  booktitle=ICCV,
  year={2025}
}

@article{zhang2025seerdrive,
  title={Future-Aware End-to-End Driving: Bidirectional Modeling of Trajectory Planning and Scene Evolution},
  author={Zhang, Bozhou and Song, Nan and Li, Jingyu and Zhu, Xiatian and Deng, Jiankang and Zhang, Li},
  journal=Neurips,
  year={2025}
}

@article{liu2024sora,
  title={Sora: A review on background, technology, limitations, and opportunities of large vision models},
  author={Liu, Yixin and Zhang, Kai and Li, Yuan and Yan, Zhiling and Gao, Chujie and Chen, Ruoxi and Yuan, Zhengqing and Huang, Yue and Sun, Hanchi and Gao, Jianfeng and others},
  journal={arXiv preprint},
  year={2024}
}

@inproceedings{wang2024driving,
  title={Driving into the future: Multiview visual forecasting and planning with world model for autonomous driving},
  author={Wang, Yuqi and He, Jiawei and Fan, Lue and Li, Hongxin and Chen, Yuntao and Zhang, Zhaoxiang},
  booktitle={CVPR},
  year={2024}
}

@inproceedings{chen2025drivinggpt,
  title={Drivinggpt: Unifying driving world modeling and planning with multi-modal autoregressive transformers},
  author={Chen, Yuntao and Wang, Yuqi and Zhang, Zhaoxiang},
  booktitle={ICCV},
  year={2025}
}

@inproceedings{zhao2025pwm,
  title={From Forecasting to Planning: Policy World Model for Collaborative State-Action Prediction},
  author={Zhao, Zhida and Fu, Talas and Wang, Yifan and Wang, Lijun and Lu, Huchuan},
  booktitle={NeurIPS},
  year={2025}
}

@article{li2025ztrs,
  title={ZTRS: Zero-Imitation End-to-end Autonomous Driving with Trajectory Scoring},
  author={Li, Zhenxin and Yao, Wenhao and Wang, Zi and Sun, Xinglong and Chen, Jingde and Chang, Nadine and Shen, Maying and Song, Jingyu and Wu, Zuxuan and Lan, Shiyi and others},
  journal={arXiv preprint},
  year={2025}
}

@article{li2025generalized,
  title={Generalized Trajectory Scoring for End-to-end Multimodal Planning},
  author={Li, Zhenxin and Yao, Wenhao and Wang, Zi and Sun, Xinglong and Chen, Joshua and Chang, Nadine and Shen, Maying and Wu, Zuxuan and Lan, Shiyi and Alvarez, Jose M},
  journal={arXiv preprint},
  year={2025}
}

@article{li2025drivevla,
  title={DriveVLA-W0: World Models Amplify Data Scaling Law in Autonomous Driving},
  author={Li, Yingyan and Shang, Shuyao and Liu, Weisong and Zhan, Bing and Wang, Haochen and Wang, Yuqi and Chen, Yuntao and Wang, Xiaoman and An, Yasong and Tang, Chufeng and others},
  journal={arXiv preprint},
  year={2025}
}

@inproceedings{shang2025drivedpo,
  title={DriveDPO: Policy Learning via Safety DPO For End-to-End Autonomous Driving},
  author={Shang, Shuyao and Chen, Yuntao and Wang, Yuqi and Li, Yingyan and Zhang, Zhaoxiang},
  booktitle={NeurIPS},
  year={2025}
}

@inproceedings{yan2025rlgf,
  title={RLGF: Reinforcement Learning with Geometric Feedback for Autonomous Driving Video Generation},
  author={Yan, Tianyi and Han, Wencheng and Zhou, Xia and Zhang, Xueyang and Zhan, Kun and Xu, Cheng-zhong and Shen, Jianbing},
  booktitle={NeurIPS},
  year={2025}
}

@inproceedings{zeng2025futuresightdrive,
  title={FutureSightDrive: Thinking Visually with Spatio-Temporal CoT for Autonomous Driving},
  author={Zeng, Shuang and Chang, Xinyuan and Xie, Mengwei and Liu, Xinran and Bai, Yifan and Pan, Zheng and Xu, Mu and Wei, Xing},
  booktitle={NeurIPS},
  year={2025}
}

@article{chi2025impromptu,
  title={Impromptu VLA: Open Weights and Open Data for Driving Vision-Language-Action Models},
  author={Chi, Haohan and Gao, Huan-ang and Liu, Ziming and Liu, Jianing and Liu, Chenyu and Li, Jinwei and Yang, Kaisen and Yu, Yangcheng and Wang, Zeda and Li, Wenyi and others},
  journal={arXiv preprint},
  year={2025}
}

@inproceedings{yang2024drivearena,
    title={DriveArena: A Closed-loop Generative Simulation Platform for Autonomous Driving}, 
    author={Xuemeng Yang and Licheng Wen and Yukai Ma and Jianbiao Mei and Xin Li and Tiantian Wei and Wenjie Lei and Daocheng Fu and Pinlong Cai and Min Dou and Botian Shi and Liang He and Yong Liu and Yu Qiao},
    booktitle={ICCV},
    year={2025}
}

@inproceedings{hamdan2025eta,
  title={ETA: Efficiency through Thinking Ahead, A Dual Approach to Self-Driving with Large Models},
  author={Hamdan, Shadi and Sima, Chonghao and Yang, Zetong and Li, Hongyang and Guney, Fatma},
  booktitle={ICCV},
  year={2025}
}

@inproceedings{zhou2025autovla,
 title={AutoVLA: A Vision-Language-Action Model for End-to-End Autonomous Driving with Adaptive Reasoning and Reinforcement Fine-Tuning},
 author={Zhou, Zewei and Cai, Tianhui and Zhao, Seth Z.and Zhang, Yun and Huang, Zhiyu and Zhou, Bolei and Ma, Jiaqi},
 booktitle={NeurIPS},
 year={2025}
}

@article{guo2025ipad,
  title={iPad: Iterative Proposal-centric End-to-End Autonomous Driving},
  author={Guo, Ke and Liu, Haochen and Wu, Xiaojun and Pan, Jia and Lv, Chen},
  journal={arXiv preprint},
  year={2025}
}

@article{chen2025rift,
  title={RIFT: Closed-Loop RL Fine-Tuning for Realistic and Controllable Traffic Simulation},
  author={Chen, Keyu and Sun, Wenchao and Cheng, Hao and Zheng, Sifa},
  journal={arXiv preprint},
  year={2025}
}

@article{ren2025cosmos,
  title={Cosmos-Drive-Dreams: Scalable Synthetic Driving Data Generation with World Foundation Models},
  author={Ren, Xuanchi and Lu, Yifan and Cao, Tianshi and Gao, Ruiyuan and Huang, Shengyu and Sabour, Amirmojtaba and Shen, Tianchang and Pfaff, Tobias and Wu, Jay Zhangjie and Chen, Runjian and others},
  journal={arXiv preprint},
  year={2025}
}

@article{jiang2025alphadrive,
  title={Alphadrive: Unleashing the power of vlms in autonomous driving via reinforcement learning and reasoning},
  author={Jiang, Bo and Chen, Shaoyu and Zhang, Qian and Liu, Wenyu and Wang, Xinggang},
  journal={arXiv preprint},
  year={2025}
}

@article{jaeger2025carl,
  title={Carl: Learning scalable planning policies with simple rewards},
  author={Jaeger, Bernhard and Dauner, Daniel and Bei{\ss}wenger, Jens and Gerstenecker, Simon and Chitta, Kashyap and Geiger, Andreas},
  journal={arXiv preprint},
  year={2025}
}

@article{sima2025centaur,
  title={Centaur: Robust end-to-end autonomous driving with test-time training},
  author={Sima, Chonghao and Chitta, Kashyap and Yu, Zhiding and Lan, Shiyi and Luo, Ping and Geiger, Andreas and Li, Hongyang and Alvarez, Jose M},
  journal={arXiv preprint},
  year={2025}
}

@article{li2025finetuning,
  title={Finetuning generative trajectory model with reinforcement learning from human feedback},
  author={Li, Derun and Ren, Jianwei and Wang, Yue and Wen, Xin and Li, Pengxiang and Xu, Leimeng and Zhan, Kun and Xia, Zhongpu and Jia, Peng and Lang, Xianpeng and others},
  journal={arXiv preprint},
  year={2025}
}

@article{jiang2025irl,
  title={Irl-vla: Training an vision-language-action policy via reward world model},
  author={Jiang, Anqing and Gao, Yu and Wang, Yiru and Sun, Zhigang and Wang, Shuo and Heng, Yuwen and Sun, Hao and Tang, Shichen and Zhu, Lijuan and Chai, Jinhao and others},
  journal={arXiv preprint},
  year={2025}
}

@article{cao2025fastdrivevla,
  title={Fastdrivevla: Efficient end-to-end driving via plug-and-play reconstruction-based token pruning},
  author={Cao, Jiajun and Zhang, Qizhe and Jia, Peidong and Zhao, Xuhui and Lan, Bo and Zhang, Xiaoan and Li, Zhuo and Wei, Xiaobao and Chen, Sixiang and Li, Liyun and others},
  journal={arXiv preprint},
  year={2025}
}

@inproceedings{chen2025solve,
  title={SOLVE: Synergy of Language-Vision and End-to-End Networks for Autonomous Driving},
  author={Chen, Xuesong and Huang, Linjiang and Ma, Tao and Fang, Rongyao and Shi, Shaoshuai and Li, Hongsheng},
  booktitle={CVPR},
  year={2025}
}

@article{tang2025hip,
  title={Hip-ad: Hierarchical and multi-granularity planning with deformable attention for autonomous driving in a single decoder},
  author={Tang, Yingqi and Xu, Zhuoran and Meng, Zhaotie and Cheng, Erkang},
  journal={arXiv preprint},
  year={2025}
}

@inproceedings{li2025uniscene,
  title={Uniscene: Unified occupancy-centric driving scene generation},
  author={Li, Bohan and Guo, Jiazhe and Liu, Hongsi and Zou, Yingshuang and Ding, Yikang and Chen, Xiwu and Zhu, Hu and Tan, Feiyang and Zhang, Chi and Wang, Tiancai and others},
  booktitle={CVPR},
  year={2025}
}

@article{wei2024occllama,
  title={Occllama: An occupancy-language-action generative world model for autonomous driving},
  author={Wei, Julong and Yuan, Shanshuai and Li, Pengfei and Hu, Qingda and Gan, Zhongxue and Ding, Wenchao},
  journal={arXiv preprint},
  year={2024}
}

@article{zhou2025hermes,
  title={Hermes: A unified self-driving world model for simultaneous 3d scene understanding and generation},
  author={Zhou, Xin and Liang, Dingkang and Tu, Sifan and Chen, Xiwu and Ding, Yikang and Zhang, Dingyuan and Tan, Feiyang and Zhao, Hengshuang and Bai, Xiang},
  journal={arXiv preprint},
  year={2025}
}

@inproceedings{yan2025drivingsphere,
  title={Drivingsphere: Building a high-fidelity 4d world for closed-loop simulation},
  author={Yan, Tianyi and Wu, Dongming and Han, Wencheng and Jiang, Junpeng and Zhou, Xia and Zhan, Kun and Xu, Cheng-zhong and Shen, Jianbing},
  booktitle={CVPR},
  year={2025}
}

@article{li2025omninwm,
  title={OmniNWM: Omniscient Driving Navigation World Models},
  author={Li, Bohan and Ma, Zhuang and Du, Dalong and Peng, Baorui and Liang, Zhujin and Liu, Zhenqiang and Ma, Chao and Jin, Yueming and Zhao, Hao and Zeng, Wenjun and others},
  journal={arXiv preprint},
  year={2025}
}

@inproceedings{gao2025magicdrive,
  title={MagicDrive-V2: High-resolution long video generation for autonomous driving with adaptive control},
  author={Gao, Ruiyuan and Chen, Kai and Xiao, Bo and Hong, Lanqing and Li, Zhenguo and Xu, Qiang},
  booktitle={ICCV},
  year={2025}
}

@inproceedings{ni2025maskgwm,
  title={Maskgwm: A generalizable driving world model with video mask reconstruction},
  author={Ni, Jingcheng and Guo, Yuxin and Liu, Yichen and Chen, Rui and Lu, Lewei and Wu, Zehuan},
  booktitle={CVPR},
  year={2025}
}

@inproceedings{yang2025driving,
  title={Driving in the occupancy world: Vision-centric 4d occupancy forecasting and planning via world models for autonomous driving},
  author={Yang, Yu and Mei, Jianbiao and Ma, Yukai and Du, Siliang and Chen, Wenqing and Qian, Yijie and Feng, Yuxiang and Liu, Yong},
  booktitle={AAAI},
  year={2025}
}

@article{jiang2025diffvla,
  title={Diffvla: Vision-language guided diffusion planning for autonomous driving},
  author={Jiang, Anqing and Gao, Yu and Sun, Zhigang and Wang, Yiru and Wang, Jijun and Chai, Jinghao and Cao, Qian and Heng, Yuweng and Jiang, Hao and Dong, Yunda and others},
  journal={arXiv preprint},
  year={2025}
}

@article{luo2025adathinkdrive,
  title={AdaThinkDrive: Adaptive Thinking via Reinforcement Learning for Autonomous Driving},
  author={Luo, Yuechen and Li, Fang and Xu, Shaoqing and Lai, Zhiyi and Yang, Lei and Chen, Qimao and Luo, Ziang and Xie, Zixun and Jiang, Shengyin and Liu, Jiaxin and others},
  journal={arXiv preprint},
  year={2025}
}

@article{jiang2025transdiffuser,
  title={TransDiffuser: End-to-end Trajectory Generation with Decorrelated Multi-modal Representation for Autonomous Driving},
  author={Jiang, Xuefeng and Ma, Yuan and Li, Pengxiang and Xu, Leimeng and Wen, Xin and Zhan, Kun and Xia, Zhongpu and Jia, Peng and Lang, XianPeng and Sun, Sheng},
  journal={arXiv preprint},
  year={2025}
}

@article{feng2025artemis,
  title={Artemis: Autoregressive end-to-end trajectory planning with mixture of experts for autonomous driving},
  author={Feng, Renju and Xi, Ning and Chu, Duanfeng and Wang, Rukang and Deng, Zejian and Wang, Anzheng and Lu, Liping and Wang, Jinxiang and Huang, Yanjun},
  journal={arXiv preprint},
  year={2025}
}

@article{li2025discrete,
  title={Discrete diffusion for reflective vision-language-action models in autonomous driving},
  author={Li, Pengxiang and Zheng, Yinan and Wang, Yue and Wang, Huimin and Zhao, Hang and Liu, Jingjing and Zhan, Xianyuan and Zhan, Kun and Lang, Xianpeng},
  journal={arXiv preprint},
  year={2025}
}

@article{jiao2025evadrive,
  title={EvaDrive: Evolutionary Adversarial Policy Optimization for End-to-End Autonomous Driving},
  author={Jiao, Siwen and Qian, Kangan and Ye, Hao and Zhong, Yang and Luo, Ziang and Jiang, Sicong and Huang, Zilin and Fang, Yangyi and Miao, Jinyu and Fu, Zheng and others},
  journal={arXiv preprint},
  year={2025}
}

@article{jiang2025flowdrive,
  title={FlowDrive: Energy Flow Field for End-to-End Autonomous Driving},
  author={Jiang, Hao and Zhang, Zhipeng and Gao, Yu and Sun, Zhigang and Wang, Yiru and Heng, Yuwen and Wang, Shuo and Chai, Jinhao and Chen, Zhuo and Zhao, Hao and others},
  journal={arXiv preprint},
  year={2025}
}

@article{zhao2025diffe2e,
  title={DiffE2E: Rethinking End-to-End Driving with a Hybrid Action Diffusion and Supervised Policy},
  author={Zhao, Rui and Fan, Yuze and Chen, Ziguo and Gao, Fei and Gao, Zhenhai},
  journal={arXiv preprint},
  year={2025}
}

@article{liu2025gaussianfusion,
  title={GaussianFusion: Gaussian-Based Multi-Sensor Fusion for End-to-End Autonomous Driving},
  author={Liu, Shuai and Liang, Quanmin and Li, Zefeng and Li, Boyang and Huang, Kai},
  journal={arXiv preprint},
  year={2025}
}
}

\setcounter{page}{1}
\maketitlesupplementary

\startcontents
{
    \hypersetup{linkcolor=black}
    \printcontents{}{1}{}
}
\newpage

\section{Discussions}
\paragraph{Discussion 1:}
\textbf{More discussions with related work.}

\noindent \textbf{World model:}

Besides the comparisons presented in the main paper, we further discuss how \netName{} differs from methods such as Epona~\cite{zhang2025epona} and DrivingGPT~\cite{chen2025drivinggpt}, highlighting the unique contributions of our approach.

Although Epona, DrivingGPT, and our model all generate future frames and plan future trajectories, their focuses are fundamentally different. Epona and DrivingGPT are primarily world-modeling methods—their core contribution lies in training generative models for future-scene synthesis, whereas trajectory planning is treated as an auxiliary output. DrivingGPT uses discrete tokens to generate future trajectories, while Epona employs a diffusion transformer for trajectory generation. Both methods jointly generate future frames and trajectories within a unified generative framework.

In contrast, our approach follows a world–action design tailored specifically for end-to-end autonomous driving. \netName{} leverages a foundation world model as a unified module for perception, comprehension, and future-scene imagination, while a specially designed action module is optimized to produce high-quality trajectories. This separation allows the system to fully exploit future predictions while maintaining strong planning performance.

\noindent \textbf{Planning model:}

Recent planning models explore diverse directions to enhance planning performance, including diffusion-based decoders~\cite{diffusiondrive,li2025recogdrive,jiang2025diffvla,jiang2025transdiffuser,li2025discrete,jiang2025flowdrive,zhao2025diffe2e}, reinforcement-learning–based approaches~\cite{luo2025adathinkdrive,jiang2025irl,chen2025rift,jiao2025evadrive,shang2025drivedpo}, test-time training~\cite{sima2025centaur}, clustered anchored trajectory priors~\cite{hydranext,li2025generalized}, mixture-of-experts architectures~\cite{feng2025artemis}, and Gaussian-feature–fusion strategies~\cite{liu2025gaussianfusion}. All of these methods demonstrate strong empirical performance.

These approaches are largely orthogonal to ours: \netName{} intentionally adopts simple action-decoding modules so that our core contribution—the integration of a foundation world model to guide planning—is clearly isolated and easy to plug into existing systems. We believe that combining these advanced planning techniques with our world–action framework would likely yield even better performance.

\paragraph{Discussion 2:}
\textbf{About the world model architecture.}

Our \netName{} relies on the generated future-frame features for planning, which means it is not restricted to any specific world-model architecture, such as diffusion-based models~\cite{zhang2025epona,vista} or GPT-based models~\cite{hu2024drivingworld,chen2025drivinggpt}. In our main experiments, we adopt Epona~\cite{zhang2025epona} as the primary world model, considering both its capability and open-source availability. Nevertheless, we also provide results using alternative world-model architectures in Section~\ref{other}, demonstrating that \netName{} is compatible with diverse backbone designs.

\begin{table} [t!]
\centering
\caption{
Generation performance comparison.}

{\begin{tabular}{c|c}
\toprule[1.5pt]
Method & FVD$_{10}$ \\
\midrule

Epona~\cite{zhang2025epona} & 50.77 \\
\rowcolor{gray!20}
\netName{}~(Ours) & 54.63 \\

\bottomrule[1.5pt]
\end{tabular}}

\label{tab:supp_gen}
\end{table}

\paragraph{Discussion 3:}
\textbf{About the current encoder.}

In autonomous driving, accurate trajectory planning often requires a detailed understanding of the surroundings, typically obtained from multi-view cameras or LiDAR point clouds. To provide this spatial awareness, we incorporate a lightweight encoder based on TransFuser~\cite{TransFuser} to extract current-frame features as an additional supplement. However, this component is not strictly necessary. As shown in Section~\ref{current}, we also report results without the current encoder. With future advances in world models—particularly in high-resolution and multi-view generation—we expect that the current encoder can eventually be removed and fully replaced by a unified world–action framework.

\paragraph{Discussion 4:}
\textbf{Efficiency analysis.}

For the parameter size, \netName{} mainly consists of three components: the foundation world model, the current encoder, and the action decoder. For the foundation world model, we adopt Epona~\cite{zhang2025epona}, which contains 2.5 B parameters. The current encoder is largely inherited from TransFuser~\cite{TransFuser}, comprising 52 M parameters. The action decoder contains an additional 21 M parameters.

For inference time, we evaluate our model on an NVIDIA H100 GPU. The average inference time of \netName{} is 900 ms, with the majority (approximately 870 ms) attributed to the world model~\cite{zhang2025epona}. As world-model architectures continue to advance, their inference efficiency is expected to improve substantially, making the overall system considerably more deployment-friendly.

\section{Experiments}

\subsection{Implementation details}
Besides the implementation details provided in the main paper, additional information is included here to ensure full reproducibility.
For the foundation world model, Epona~\cite{zhang2025epona} is adopted. Its native generation frequency is 5 Hz, whereas the planning frequency in our system is 2 Hz. To match this temporal resolution, Epona is first finetuned on the nuPlan~\cite{nuplan} dataset at 2 Hz and subsequently frozen when training the full pipeline on NAVSIM~\cite{navsim}.
For the current encoder, the design largely follows TransFuser~\cite{TransFuser}.
During training, the current encoder and the action decoder are first pretrained without the future feature cross-attention or the WM-QFormer modules.
Afterwards, the full model is trained end-to-end with all components enabled, except that the world model remains frozen.

\begin{table} [t!]
\centering
\caption{
Performance comparison with and without the current encoder.}

{\begin{tabular}{c|cccc}
\toprule[1.5pt]
 & DAC $\uparrow$ & TTC $\uparrow$ & EP $\uparrow$ &  PDMS $\uparrow$  \\
\midrule
w/o Current & 96.3 & 95.4 & 81.7 & \cellcolor{gray!20}88.2 \\
\rowcolor{gray!20}
\netName{} & 97.2 & 94.8 & 83.5 & \cellcolor{gray!20}89.3  \\

\bottomrule[1.5pt]
\end{tabular}}

\label{tab:supp_curr}
\end{table}

\begin{table*} [ht!] 
\centering
\caption{
Performance with different world-model architectures on the nuScenes dataset for the planning task.
}
{\begin{tabular}{l|cccc|cccc}
\toprule[1.5pt]
\multirow{2}{*}{Method} &
\multicolumn{4}{c|}{L2 ($m$) $\downarrow$} & 
\multicolumn{4}{c}{Col. Rate (\%) $\downarrow$} \\
& 1$s$ & 2$s$ & 3$s$ & Avg. & 1$s$ & 2$s$ & 3$s$ & Avg. \\
\midrule

\netName-Vista                 & 0.42 & 0.63 & 0.88 & 0.64 & 0.08 & 0.22 & 0.51 & 0.27 \\
\netName-Epona                 & 0.36 & 0.55 & 0.93 & 0.62 & 0.04 & 0.12 & 0.37 & 0.18 \\

\bottomrule[1.5pt]
\end{tabular}}


\label{tab:supp_vista}
\end{table*}

\subsection{More experiment results}

\paragraph{Generation performance of the world model.}
As shown in Table~\ref{tab:supp_gen}, we report the Fréchet Video Distance (FVD) of \netName{} and Epona~\cite{zhang2025epona} on the nuPlan~\cite{nuplan} dataset. The results indicate that \netName{} retains nearly the same generation capability as Epona after finetuning.

\paragraph{Performance without the current encoder.}
\label{current}
As shown in Table~\ref{tab:supp_curr}, we report the performance of our model without the current encoder on the NAVSIM~\cite{navsim} dataset. The results show that the model still achieves strong performance even when this module is removed, demonstrating that it is not strictly necessary. Nevertheless, the current encoder is retained in the full system, as it further enhances robustness and overall capability.

\paragraph{Performance with an alternative world-model architecture.}
\label{other}
As shown in Table~\ref{tab:supp_vista}, we also evaluate our framework using Vista~\cite{vista} as the foundation world model. Since Vista is trained on the nuScenes~\cite{nuscenes} dataset, the experiments are conducted on this dataset for the end-to-end planning task. The results indicate that \netName{} with Vista achieves strong performance as well, demonstrating that our framework is not restricted to a specific world-model architecture.

\section{Qualitative results}
\begin{figure*}[t!]
\centering
\includegraphics[width=1\textwidth]{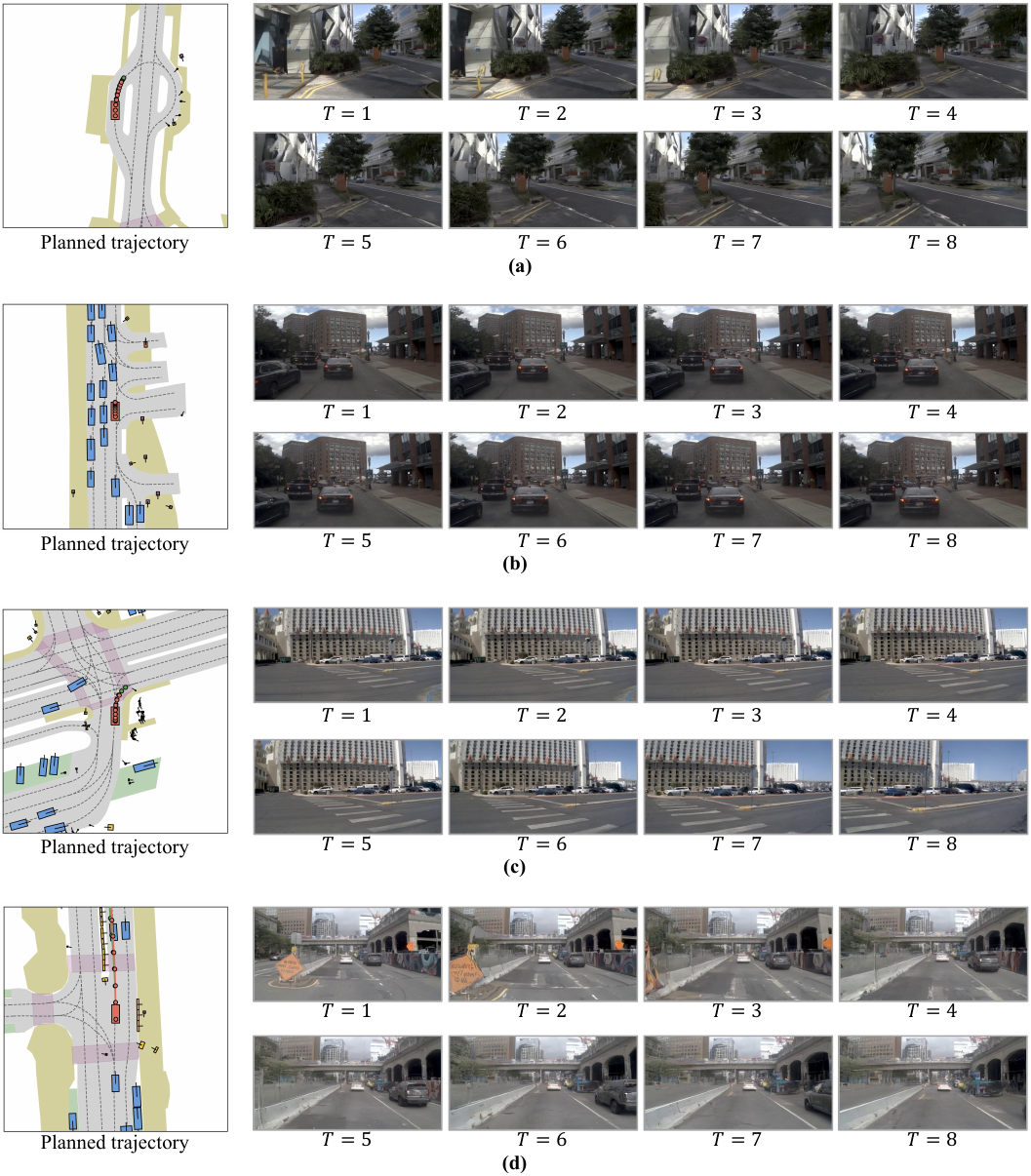}
\caption{
Visualization of \textbf{\netName{}} on the NAVSIM~\cite{navsim} dataset. The left panel shows the planned trajectories in the BEV view, while the right panel presents the generated future video over the next 8 time steps. The ground-truth trajectory is depicted in \textcolor{ForestGreen}{green}, and the final planned trajectory is highlighted in \textcolor{orange}{orange}.
}
\label{fig:visual_supp}
\end{figure*}

As shown in Figure~\ref{fig:visual_supp}, additional qualitative results of \netName{} are provided, including turning behaviors (parts (a) and (c)), a traffic-congestion scenario (part (b)), and a fast-driving behavior (part (d)). Across all cases, the model not only predicts future frames accurately but also produces precise future trajectories.

\section{Failure cases}
\begin{figure*}[t!]
\centering
\includegraphics[width=1\textwidth]{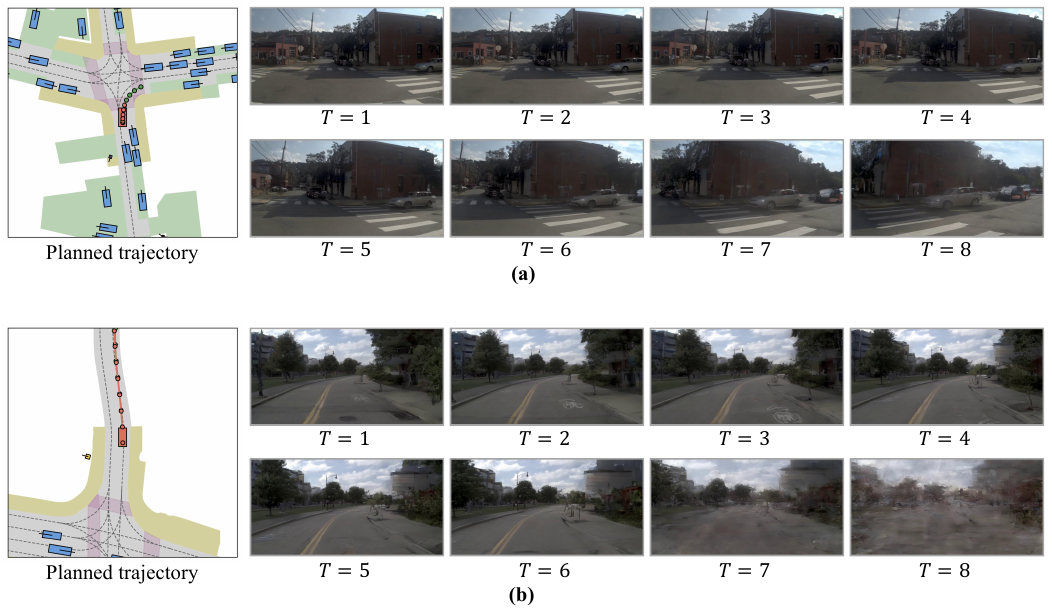}
\caption{
Visualization of \textit{failure cases} for \textbf{\netName{}} on the NAVSIM~\cite{navsim} dataset. The left panel shows the planned trajectories in the BEV view, while the right panel presents the generated future video over the next 8 time steps. The ground-truth trajectory is depicted in \textcolor{ForestGreen}{green}, and the final planned trajectory is highlighted in \textcolor{orange}{orange}.
}
\label{fig:visual_fail}
\end{figure*}

Although \netName{} is powerful, it still fails in certain scenarios. Representative failure cases are shown in Figure~\ref{fig:visual_fail}, which may provide insights for future research.

In part (a), the scenario involves a right-turning maneuver. The foundation world model accurately predicts both the turning motion and the post-turn scene. However, the action decoder generates an overly conservative and slow trajectory. This indicates that the world model and the action model should be more tightly coupled so that the planner can better leverage future predictions for trajectory generation.

In part (b), the scenario involves fast driving over a long distance within the planning horizon, and the road is highly winding. While the foundation world model produces accurate predictions initially, it fails in the later stages due to the increasing curvature of the road. This suggests that long-range prediction capability remains a challenge for the world model. Nevertheless, the action model still produces an accurate trajectory. This highlights the importance of using current-frame features as an additional supplement, which significantly enhances the overall robustness of the system.

\end{document}